\documentclass[letterpaper, 10 pt, conference]{ieeeconf}  
\IEEEoverridecommandlockouts                              
\usepackage{cite}
\usepackage{todonotes}
\usepackage[nolist,nohyperlinks]{acronym}
\usepackage[super]{nth}
\usepackage{amsmath,amssymb,amsfonts,mathtools}
\usepackage{algorithmic}
\usepackage{graphicx}  
\usepackage{subcaption} 
\usepackage{textcomp} 
\usepackage{xcolor}
\usepackage[protrusion=true,
             expansion=true,
             tracking=true,
             final,
             stretch=50,factor=1350]{microtype}

\mathchardef\mhyphen="2D

\def\ie{\emph{i.e}.\ }

\def\etal{\emph{et al}.\ }

\DeclareRobustCommand{\mysmallmath}[1]{\scriptscriptstyle{#1}}

\newlength{\depthofsumsign}
\newcommand{\nsum}[2][1.4]{
    \mathop{%
        \raisebox
            {-#1\depthofsumsign+1\depthofsumsign}
            {\scalebox
                {#1}
                {$\displaystyle\sum_i^{\mysmallmath{#2}}$}%
            }
    }
}

\newcommand{\brackets}[3]{\ensuremath{\left#1 #2 \right#3}}
\newcommand{\mypar}[1]{\brackets{(}{#1}{)}}
\newcommand{\mybra}[1]{\brackets{\{}{#1}{\}}}

\newcommand{\idx}[2][]{\ifthenelse{\equal{#1}{}}{#2}{#2{#1}}}
\newcommand{\mytime}[1][]{\idx[#1]{t}}

\newcommand{\mysym}[5]{\ensuremath{\prescript{#1}{#2\hspace{+0.25mm}}{#3}_{#4}^{#5}}}

\newcommand{\mysca}[1]{\ensuremath{\lowercase{#1}}}
\newcommand{\myvtr}[1]{\ensuremath{\mathbf{\lowercase{#1}}}}

\newcommand{\myset}[1]{\ensuremath{\mathbb{\MakeUppercase{#1}}}}
\newcommand{\mymat}[1]{\ensuremath{\mathbf{\uppercase{#1}}}}
\newcommand{\myfunc}[1]{\ensuremath{\boldsymbol{\lowercase{#1}}}}

\newcommand{\sca}[5]{\mysym{#1}{#2}{\mysca{#3}}{#4}{#5}}
\newcommand{\easysca}[3]{\mysym{}{}{\mysca{#1}}{#2}{#3}}
\newcommand{\vtr}[5]{\mysym{#1}{#2}{\myvtr{#3}}{#4}{#5}}

\newcommand{\set}[5]{\mysym{#1}{#2}{\myset{#3}}{#4}{#5}}
\newcommand{\mat}[5]{\mysym{#1}{#2}{\mymat{#3}}{#4}{#5}}
\newcommand{\fnc}[5]{\mysym{#1}{#2}{\myfunc{#3}}{#4}{#5}}
\newcommand{\func}[6]{\fnc{#1}{#2}{#3}{#4}{#5}\mypar{#6}}
\newcommand{\easyfunc}[2]{\func{}{}{#1}{}{}{#2}}

\newcommand{\lss}[4]{\mat{#1}{#2}{\mathcal{L}}{#3}{#4}}

\newcommand{\mytuple}[1]{\ensuremath{\left\langle #1 \right\rangle}}

\newcommand{\myprob}[2]{\ensuremath{\Pr_{#2}\mypar{#1}}}

\newcommand{\myloss}[1]{\lss{}{}{#1}{}}

\newcommand{\psn}[0]{x}

\newcommand{\pos}[0]{\dot{\psn}}
\DeclareRobustCommand{\pose}[2]{\vtr{}{}{\pos}{#1}{#2}}
\DeclareRobustCommand{\poseSet}[2]{\set{}{}{\pos}{#1}{#2}}

\newcommand{\stt}[0]{p}
\newcommand{\state}[4]{\vtr{#1}{#2}{\stt}{#3}{#4}}
\newcommand{\statespace}[4]{\set{#1}{#2}{\stt}{#3}{#4}}

\newcommand{\os}[0]{z}
\newcommand{\obs}	[2]	{\mysym{}{}{\os}{#1}{#2}}
\newcommand{\obsall}	[2]	{\set{}{}{\os}{#1}{#2}}
\newcommand{\om}[0]{u}
\newcommand{\odom}	[2]	{\mysym{}{}{\om}{#1}{#2}}

\newcommand{\odomall}	[2]	{\set{}{}{\om}{#1}{#2}}

\newcommand{\condprob}[2]{\ensuremath{\myprob{ #1 \middle| #2 }{}}}

\begin{document}
%
%
%
\begin{acronym}[RANSAC] 
\acro{nbv}[NBV]{Next-Best View}
\acro{nbs}[NBS]{Next-Best Stereo}

\acro{bmvc}[BMVC]{British Machine Vision Conference}
\acro{iccv}[ICCV]{International Conference on Computer Vision}

\acro{agast}[AGAST]{Adaptive and Generic Accelerated Segment Test}
\acro{fast}[FAST]{Features from Accelerated Segment Test}

\acro{ate}[ATE]{Absolute Trajectory Error}
\acro{rpe}[RPE]{Relative Pose Error}
\acro{rmse}[RMSE]{Root Mean Square Error}

\acro{dof}[DoF]{Degree of Freedom}
\acro{se2}[SE(2)]{Special Euclidean Space}
\acro{so2}[SO(2)]{Special Orthogonal Space}
\acro{se3}[SE(3)]{Special Euclidean Space}
\acro{so3}[SO(3)]{Special Orthogonal Space}

\acro{cnn}[CNN]{Convolutional Neural Network}
\acro{fcn}[FCN]{Fully Convolutional Network}
\acro{lut}[LUT]{Lookup Table}
\acro{ls}[LS]{Least-Squares}
\acro{kdt}[k-d tree]{k-dimensional Tree}
\acro{if}[IF]{Information Filter}
\acro{lls}[L-LS]{Linear Least-Squares}
\acro{ills}[ILLS]{Iterative Linear-Least-Squares}
\acro{mvs}[MVS]{Multi-View Stereo}
\acro{mocap}[MoCap]{Motion Capture}
\acro{pid}[PID]{Proportional Integral Controller}
\acro{vp}[VP]{Vanishing Point}
\acro{svd}[SVD]{Single Value Decomposition}
\acro{ba}[BA]{Bundle Adjustment}
\acro{sf}[SF]{Sensor-Fusion}
\acro{sfm}[SfM]{Structure from Motion}
\acro{csfm}[CSfM]{Collaborative Structure from Motion}
\acro{vo}[VO]{Visual Odometry}
\acro{mse}[MSE]{Mean Squared Error}

\acro{klf}[KF]{Kalman Filter}
\acro{ekf}[EKF]{Extended Kalman Filter}
\acro{ukf}[UKF]{Unscented Kalman Filter}

  \acro{icp}[ICP]{Iterative Closest Point}
  \acro{lsd}[LSD]{Line Segment Detector}
  \acro{pnp}[PnP]{Perspective N-Points}
  \acro{rsc}[RANSAC]{RANdom SAmple and Consensus}
  \acro{svo}[SVO]{Semi-Direct Visual Odometry}
  \acro{vsfm}[VSFM]{Visual Structure from Motion}
  
  \acro{slam}[SLAM]{Simultaneous Localisation and Mapping}
  \acro{vslam}[VSLAM]{Vision-Based Simultaneous Localisation and Mapping}
  
  \acro{etam}[ETAM]{ETHZASL-PTAM}
  \acro{ptam}[PTAM]{Parallel Tracking and Mapping}
  \acro{dtam}[DTAM]{Dense Tracking and Mapping}

  \acro{pf}[PF]{Particle Filter}
  \acro{rbpf}[RBPF]{Rao-Blackwellised Particle Filter}
  \acro{smc}[SMC]{Sequential Monte-Carlo}
  \acro{mkl}[MKL]{Markov Localisation}
  \acro{gbl}[GBL]{Grid-Based Localisation}
  \acro{mcl}[MCL]{Monte-Carlo Localisation}
  \acro{amcl}[AMCL]{Adaptive Monte Carlo Localisation}
  \acro{vmcl}[VMCL]{Vision-Based Monte-Carlo Localisation}
  \acro{rmcl}[RMCL]{Range-Based Monte-Carlo Localisation}
  
  \acro{prm}[PRM]{Probabilistic Road Map}
  \acro{rrt}[RRT]{Rapidly-exploring Random Tree}
  \acro{rrt*}[RRT*]{Rapidly-exploring Random Tree}

\acro{ompl}[OMPL]{Open Motion Planning Libraries}
\acro{pcl}[PCL]{Point Cloud Library}
\acro{ros}[ROS]{Robot Operating System}

\acro{gps}[GPS]{Global Positioning System}
\acro{imu}[IMU]{Inertial Measurement Unit}
\acro{lidar}[LiDAR]{Light Detection And Ranging}
\acro{rgb}[RGB]{Red, Green and Blue}
\acro{rgbd}[RGB-D]{\acs{rgb} and Depth}
\acro{rgbds}[RGB-DS]{\acs{rgb}, Depth and Semantic}
\acro{sedar}[SeDAR]{Semantic Detection and Ranging}
\acro{smcl}[SeMCL]{Semantic Monte-Carlo Localisation}
\acro{sonar}[SoNAR]{Sound Navigation And Ranging}

\acro{ada}[ADA]{ARDrone Autonomy}
\acro{ai}[AI]{Artificial Intelligence}
\acro{ar}[AR]{Augmented Reality}
\acro{ard}[ARD]{AR Drone}
\acro{mct}[MTC]{MATLAB Calibration Toolbox}
\acro{tma}[TMA]{TUM\_ARDrone}
\acro{tum}[TUM]{Technical University of Munich}
\acro{uav}[UAV]{Unmanned Aerial Vehicle}
\acro{sdk}[SDK]{Software Development Kit}
\acro{fps}[FPS]{Frames Per Second}
\end{acronym}

\title{\LARGE \bf Markov Localisation using Heatmap Regression and Deep Convolutional Odometry} 
\author{Oscar Mendez$^{1}$, Simon Hadfield$^{1}$, Richard Bowden$^{1}$
\thanks{$^{1}$ Centre for Vision Speech and Signal Processing, University of Surrey, Guildford, UK \{o.mendez, s.hadfield, r.bowden\}@surrey.ac.uk \newline
$*$Funded  by  InnovateUK Autonomous Valet Parking Project Grant No 104273.
}}

\maketitle

\begin{abstract}
In the context of self-driving vehicles there is strong competition between approaches based on visual localisation and \ac{lidar}.
While \ac{lidar} provides important depth information, it is sparse in resolution and expensive. On the other hand, cameras are low-cost and recent developments in deep learning mean they can provide high localisation performance. However, several fundamental problems remain, particularly in the domain of uncertainty, where learning based approaches can be notoriously over-confident. 

Markov, or grid-based, localisation was an early solution to the localisation problem but fell out of favour due to its computational complexity. Representing the likelihood field as a grid (or volume) means there is a trade off between accuracy and memory size. Furthermore, it is necessary to perform expensive convolutions across the entire likelihood volume. Despite the benefit of simultaneously maintaining a likelihood for all possible locations, grid based approaches were superseded by more efficient particle filters and Monte Carlo sampling (MCL). However, MCL introduces its own problems e.g. particle deprivation. 

Recent advances in deep learning hardware allow large likelihood volumes to be stored directly on the GPU, along with the hardware necessary to efficiently perform GPU-bound 3D convolutions and this obviates many of the disadvantages of grid based methods. In this work, we present a novel \acs{cnn}-based localisation approach that can leverage modern deep learning hardware.
By implementing a grid-based Markov localisation approach directly on the GPU, we create a hybrid \ac{cnn} that can perform image-based localisation and odometry-based likelihood propagation within a single neural network.
The resulting approach is capable of outperforming direct pose regression methods as well as state-of-the-art localisation systems.
\end{abstract}

\section{Introduction}

The reasoning that humans can localise using vision alone, has been used extensively to motivate machine localisation from visual sensors such as cameras.
However, there has always been a significant gap in the performance obtained from vision compared to \ac{lidar} and/or \ac{gps}.
Recent advances in vision use Deep-Learning based localisation \cite{Kendall2015Posenet,Mendez18} to bridge this gap by
employing \acp{cnn} to regress the camera pose directly from images.
The network learns an implicit mapping between scene appearance and location. 
However, the mapping cannot generalise beyond the training data and due to the one-to-one mapping, provides uni-modal estimates in the pose-likelihood space \cite{kendall2016modelling}.

Most traditional sampling-based localisation approaches, such as Markov/Grid-based localisation or more modern \ac{mcl}, are based on the idea that maintaining multiple hypothesis is an important part of the localisation problem.
This makes problems like global localisation (kidnapped robot) more stable.
It also allows algorithms to deal with self-similarity in environments.

Grid based methods model the likelihood of every location in the map as a set of discrete states. The resolution of the grid therefore affects accuracy. This results in a trade-off between accuracy and memory where larger grids are slower to process. To solve this problem and scale to large spaces, \ac{mcl} was proposed. \ac{mcl} samples the space using a \ac{pf}.
This makes the process computationally efficient, but such approaches suffer from particle depletion, non-uniform sampling, sample size tuning and poor parallelisation.

This work proposes a novel deep learning architecture that maps a single image into a pose-likelihood.
The network incorporates a grid based markov localisation framework to estimate a robot's pose.
To make this tractable and overcome the limitations that gave rise to \ac{mcl}, we introduce a first-of-its-kind convolutional likelihood propagation approach that models each odometry update as a single call to a gpu-bound convolution operation as part of the neural network.
This hybrid \ac{cnn} allows us to leverage the advances of deep-learning hardware to make grid-based localisation tractable by developing a single CNN architecture that can perform pose regression, localisation and odometry based likelihood propagation, in a single network, efficiently and with one forward pass.
\vspace{-0.25cm}
\section{Related Work}
One of the first implementations of a localisation\footnote{While \ac{slam} can be used as a method to solve the localisation problem \cite{Caselitz2016, Chu2016}, here we focus on approaches that explicitly tackle localisation within a known-environment} algorithm was \ac{ekf} Localisation \cite{Thrun2006ekf}.
This approach suffered from many limitations, but the most fundamental of which is the fact that it assumes that the localisation likelihood is a uni-modal distribution.
This is such a crucial limitation, that grid-based localisation \cite{simmons1995probabilistic, burgard1996estimating,lim2002grid} algorithms quickly replaced it as the state-of-the-art.
However, the computational efficiency of grid-based localisation algorithms limited their application.
As a response to this, \ac{mcl} algorithms \cite{Fox1999a,Dellaert1999} became the de-facto standard for localisation.
More recently, approaches in autonomous agent localisation have leveraged advances in deep learning.
One of the most common family of approaches is PoseNet \cite{Kendall2015Posenet} and its derivatives \cite{brahmbhatt2018geometry, Henriques2018mapnet, naseer2017deep, Walch17}.
Fundamentally, these approaches rely on sensor/pose pairs to train a \ac{cnn} that can regress poses given an input sensor measurement. 
Kendall \etal \cite{Kendall2015Posenet} first introduced the method of regressing pose from images by first encoding the images using an encoder network. 
This maps the image into a lower-dimensional latent space that can then be mapped to a 6-\ac{dof} pose using a series of fully connected layers. 
In subsequent work, Kendall \cite{kendall2017geometric} introduced a better geometric loss function that allowed faster convergence and performance and in \cite{kendall2016modelling} began to model the underlying Bayesian statistics of the localisation problem.
The nature of the networks mean a uni-modal distribution for the pose estimate is provided \ie the network regresses a single location for any given input.
However, it is well understood that regardless of the sensor used, there will often be areas of self-similarity in the environment. In this work, we regress a \textit{pose-likelihood heatmap} which provides multi-modal distributions across pose.

Sattler \etal \cite{sattler2019understanding} showed the limitations of PoseNet-like models.
They demonstrated that models are only reliable at approximating an agent's pose at a series of \textit{base poses} and do not generalise to unseen poses between these bases.
They concluded that PoseNet-like models are only reliable for coarse pose estimation and this calls into question the reliability of learning-based methods that do not employ other sources of information.
For example, temporal accumulation and odometry \cite{brahmbhatt2018geometry}.
Yang \etal \cite{yang2020d3vo} combined a PoseNet-like architecture with a depth estimation network in order to estimate motion and uncertainty.
However, they only model uncertainty in their depth estimates.

We combine modern deep learning with tried and tested approaches to motion propagation. P\"{o}schmann \etal \cite{Poschmann2017} and Mendez \etal \cite{Mendez18} both combined deep-learning segmentation with \ac{mcl} to provide a robust localisation approach.
Similarly, Neubert \etal \cite{Neubert2017} used depth regression with \ac{mcl} and \ac{icp}.
Our pose-likelihood heatmaps are a \ac{cnn}-based sensor model combined with an odometry source.
However, we implement this as a single hybrid \ac{cnn} with all operations performed on a single GPU in one forward pass making it extremely efficient. 

Although superseded by \ac{mcl}, Grid-based localisation has some important advantages e.g. not suffering from particle deprivation, robustness to ``kidnapped robot'', lack of expensive re-sampling operations and generally being well-suited to massive parallelisation.
However, grid-based localisation approaches have traditionally struggled to maintain robust estimates of pose due to the computational complexity of estimating sensor and motion models for each cell in the grid. Coarse quantisation is typically employed to make the approach tractable.
There are several methods to improve the performance of grid-based localisation, such as \cite{burgard1999experiences}.
However, they rely on less frequent sensing and/or motion integration.
We overcome these limitations by combining the pose estimation and motion model into a single neural network that can make efficient use of the GPU. 

\begin{figure*}[t]
    \centering
    \includegraphics[width=0.8\linewidth]{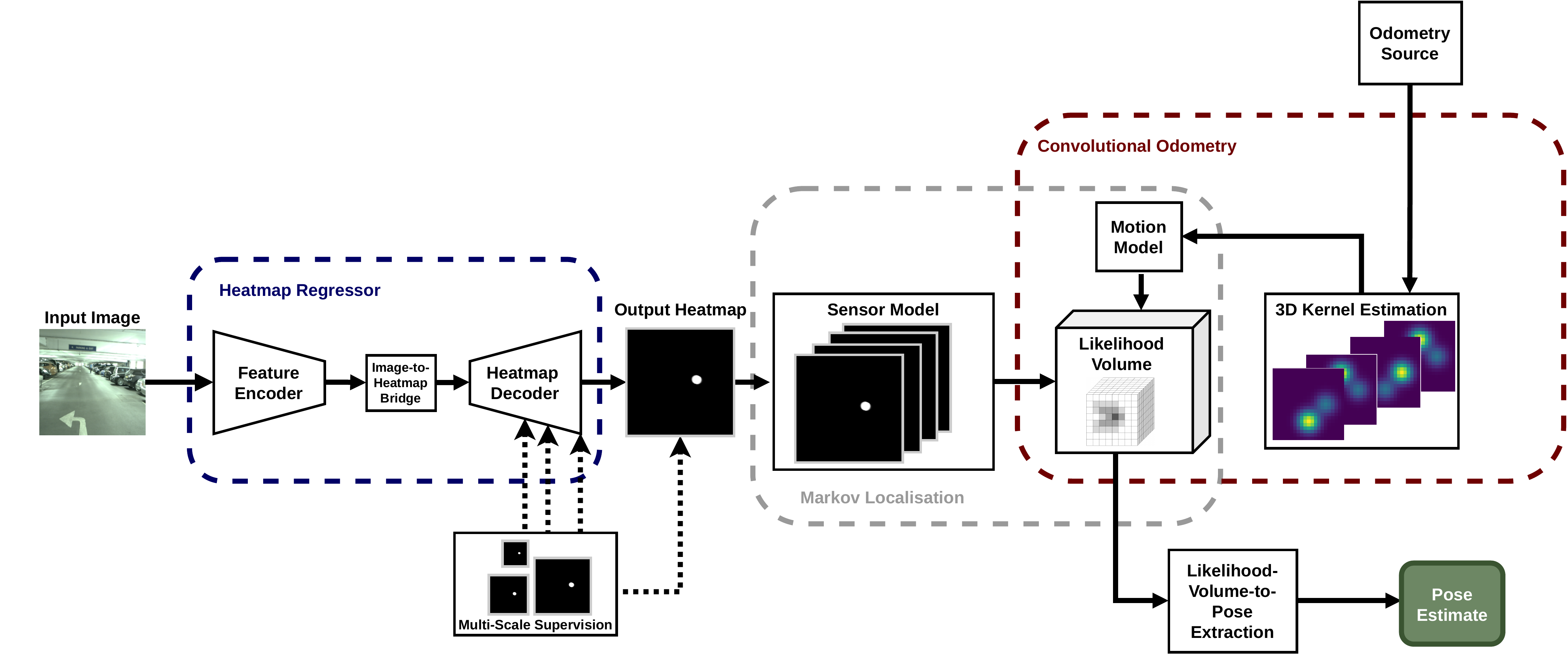}  
    \caption{Deep Markov Localisation: System Diagram.}
    \label{fig:arch}
    \vspace{-0.5cm}
\end{figure*}

\section{Methodology}
We propose to leverage the advances of deep learning to make Markov Localisation not only tractable, but also gain state-of-the-art performance.
We do this by introducing a novel hybrid \ac{cnn} architecture that combines a feature encoder layer, a image-to-heatmap-feature bridge, a heatmap decoder layer with multi-level supervision and finally a convolutional odometry layer.
Fig. \ref{fig:arch} shows an overview of the architecture used for our hybrid \ac{cnn}. Note, this entire pipeline resides on the GPU as a single network allowing both image regression, localisation and odometry updates to be done in a single forward pass. 

\subsection{Markov Localisation} \label{markov-math}
Markov localisation, shown by the gray dotted line on fig. \ref{fig:arch}, operates by taking the state-space of the autonomous agent, given by 
\ensuremath{\state{}{}{\mytime}{} \in \statespace{}{}{\mytime}{}}
and discretising it into in a grid defined as
\begin{equation}
\pose{\mytime}{k} \in \poseSet{\mytime}{}
\end{equation} 
where each \ensuremath{\pose{\mytime}{k}} is a cell in the grid \ensuremath{\poseSet{\mytime}{}} at time \ensuremath{\mytime}.
The grid \ensuremath{\poseSet{\mytime}{}} spans all possible states in the state-space \ensuremath{\statespace{}{}{\mytime}{}}.
For a ground-based vehicle, it is sufficient to represent the state space of the vehicle as a 3-\acs{dof} vector \ensuremath{\pose{}{} = \mytuple{x, y, \theta}}.
This means that the discretised grid \ensuremath{\poseSet{\mytime}{}} is a 3 dimensional volume.
This volume consists of \ensuremath{\mytuple{x,y}} planes and \ensuremath{\theta} slices.
More explicitly, this volume consists of a tensor of size \ensuremath{[\Theta \times X \times Y]}, where each cell represents the likelihood that the robot's position lies within that cell's bounds.
Under a Markov assumption, this likelihood can be defined as
\begin{equation}
\begin{split}
  \condprob{\pose{\mytime}{k}}{\obsall{\mytime}{}, \odomall{\mytime}{}} = \\
  \condprob{\obs{\mytime}{}}{\pose{\mytime}{k}} \condprob{\pose{\mytime}{k}}{\odom{\mytime}{}, \pose{\mytime-1}{k}} \condprob{\pose{\mytime-1}{k}}{\obsall{\mytime-1}{}, \odomall{\mytime-1}{}}
\end{split}
\end{equation}
which implies that the pose likelihood is conditioned on the sensor observations \ensuremath{\obs{\mytime}{} \in \obsall{\mytime}{}} and the odometry measurements \ensuremath{\odom{\mytime}{} \in \odomall{\mytime}{}} and is fully described by the sensor model
\begin{equation}
\label{sensormodel}
  \condprob{\obs{\mytime}{}}{\pose{\mytime}{k}}
\end{equation}
the motion model
\begin{equation}
\label{motionmodel}
  \condprob{\pose{\mytime}{k}}{\odom{\mytime}{}, \pose{\mytime-1}{k}}
\end{equation}
and the prior likelihood of the pose
\begin{equation}
\condprob{\pose{\mytime-1}{k}}{\obsall{\mytime-1}{}, \odomall{\mytime-1}{}}
\end{equation} 
which implies this measurement can be performed iteratively.
In this work, we use a heatmap regression sensor model (\ref{pose-heatmap}) to update the \ensuremath{[X \times Y]} volume, while using convolutional odometry (\ref{conv-odom}) for the orientation.
We additionally perform a likelihood-volume-to-pose extraction, which consists of fitting a Gaussian distribution to the likelihood volume and reporting the mean.
\subsection{Deep Likelihood Heatmap Regressor}\label{pose-heatmap} 
One of the main reasons grid-based localisation was widely considered intractable was because the sensor model (equation \ref{sensormodel}) has to be estimated for every cell in the grid.
Estimating this likelihood for sensors such as \acs{lidar}, \acs{sonar} or even RGB-D cameras can involve expensive ray-casting for each beam.
This presents a practical issue when there are large areas of the grid that we are almost certain \textit{not} to be the correct location.
Instead, we propose to use a \ac{fcn} trained for pose regression to replace the sensor model.

Using an \ac{fcn} has several important advantages.
Firstly, it allows a sensor model to be trained for any arbitrary sensor without the need for explicit mathematical derivation (although requiring training data).
Secondly, the likelihood for every cell can be estimated simultaneously.

\subsubsection{Feature Encoder Layer}

To regress a likelihood for all cells, we use an encoder-decoder architecture. 
The architecture is shown in the blue dotted line on fig. \ref{fig:arch}.
We use a ResNet encoder arm \cite{he2016deep} similar to PoseNet \cite{kendall2016modelling} but rather than a fully connected layer to regress pose, we use the decoder arm of the network (discussed below) to force the network to learn a birds-eye-view likelihood map.
We modify the ResNet by removing the fully connected layers and replace them with a convolutional image-to-heatmap bridge to the decoder. The whole network is trained end-to-end.

\subsubsection{Heatmap Decoder Layer}\label{sec:hdl}
The decoder is composed of a series of upsample blocks, which scale the image up by a factor of two.
Each upsample block contains a deconvolution followed by two convolution blocks.
The deconvolution is performed with a \ensuremath{[2\times2]} kernel and a stride of \ensuremath{2}.
The convolution blocks consist of a \ensuremath{[3\times3]} convolution with a stride of \ensuremath{1} followed by batch normalisation and ReLU.
The output block additionally contains a final convolution layer with \ensuremath{[1\times1]} kernel and a stride of \ensuremath{1} ensuring the desired number of output channels is achieved.
The result is a map of size \ensuremath{[N \times M \times M]} where \ensuremath{M} depends on the number of upsample blocks and \ensuremath{N} is the number of channels.

Intuitively, we could map this \ensuremath{[N \times M \times M]} output volume directly to the grid we are localising in.
In this mode, the volume would represent \ensuremath{[\Theta \times X \times Y]} in each channel respectively.
This ties the spatial resolution to the size of each channel, and the angular resolution to the number of channels.
Traditional Markov localisation estimates the sensor model this way because the ray-casting operations need to be performed differently for each \ensuremath{\theta} bin.
However, this is not an optimal use of the network as it can directly estimate the probability of an \ensuremath{\pose{\mytime}{k}} cell without reasoning about the orientation.
Furthermore, using the output volume this way would force the network to grow dramatically as the angular resolution increases.
This not only increases the number of parameters, but actually defines a very complicated regression problem.
Instead, we use each output channel as a ``likelihood band'' which allows us to treat the heatmap regression as a classification problem.

The likelihood volume represents \ensuremath{\condprob{\obs{\mytime}{}}{\pose{\mytime}{k}}} discretised into \ensuremath{[N \times M \times M]} bins where \ensuremath{N} represents the number of likelihood bands, \ensuremath{M} denotes the \ensuremath{x} and \ensuremath{y} resolution and \ensuremath{k} spans each of the pixels on the \ensuremath{XY} plane across all orientations.
More explicitly, 
\begin{equation}
    \condprob{\obs{\mytime}{}}{\pose{\mytime}{k}} \propto \condprob{\easysca{n}{t}{}}{\pose{\mytime}{k}}
    \label{eq:cla}
\end{equation}
where \ensuremath{\easysca{n}{t}{} \in N} is the likelihood band of \ensuremath{\pose{\mytime}{k}} at time \ensuremath{\mytime}.
This discretisation of the probability space allows us to treat likelihood regression as a classification problem, where we classify each \ensuremath{\pose{\mytime}{k}} cell into a pose likelihood bin.

\subsubsection{Multi-Scale Supervision}
Using the regression-to-classification mapping defined in equation \ref{eq:cla}, we can define a cross entropy loss function (after softmax)
\begin{equation}
    \myloss{c}(\easysca{n}{t}{}, \pose{\mytime}{k}) = -\condprob{\easysca{n}{t}{}}{\pose{\mytime}{k}} \cdot \log\mypar{\mathcal{R}(I)_{x,y}},
\end{equation}
where \ensuremath{R} is our heatmap regressor network, \ensuremath{I} is the input image and \ensuremath{x,y} denote the pixel location in the resulting heatmap.
We additionally provide supervision in the form of an \acs{mse} loss, defined as
\begin{equation}
    \myloss{m} = \left|\left| \condprob{\easysca{n}{t}{}}{\pose{\mytime}{k}} - \mathcal{R}(I)_{x,y} \right|\right|_{L_{2}}.
\end{equation}
The resulting loss function is defined as a weighted combination of these losses,
\begin{equation}
    \myloss{} = \myloss{c} + \omega \myloss{m}.
\end{equation}
where \ensuremath{\omega} is a hyperparameter.
Empirically, we have found that \ensuremath{\omega = 0.1} ensures the network can produce accurate heatmaps with a smooth distribution around the correct cell.

\begin{figure}
    \centering
    \includegraphics[width=0.8\linewidth]{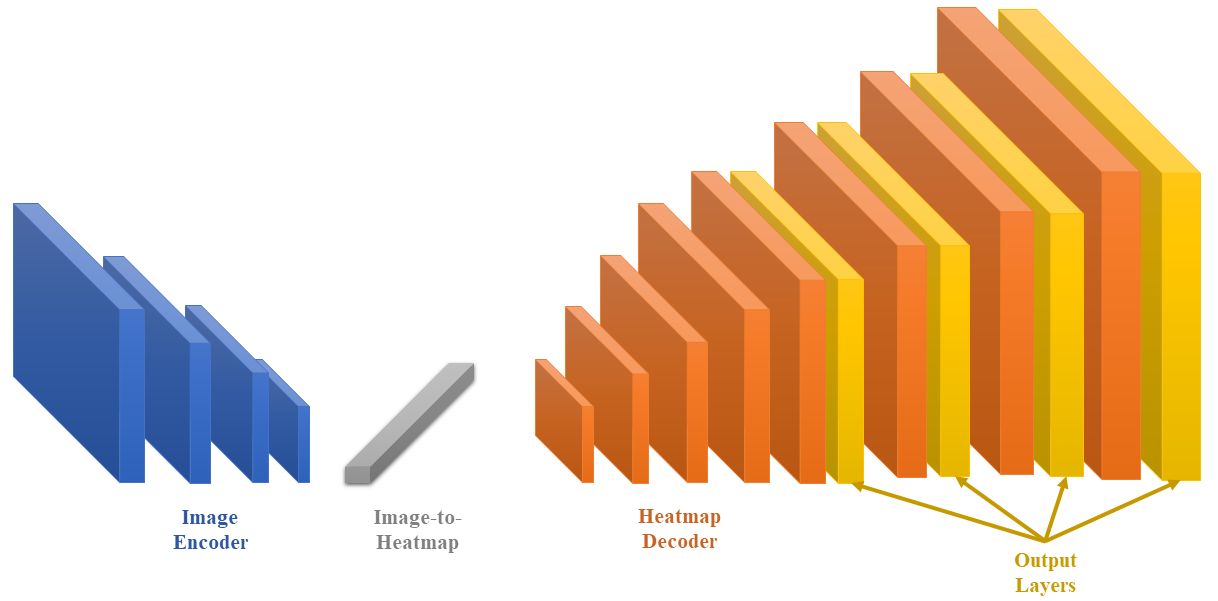}  
    \caption{Deep Markov Localisation: System Diagram.}
    \vspace{-0.5cm}
    \label{fig:reg} 
\end{figure}
To ensure a more robust loss, we perform this loss function over several different scales.
Fig. \ref{fig:reg} shows the proposed architecture, which consists of 4 encoder residual blocks (blue), 4 base heatmap decoder blocks (orange), followed by 4 output decoder blocks (orange + yellow). Note that the yellow output layers do not feed into the upscaled orange blocks, but rather directly produce an output.
Each of these output heatmaps can be directly supervised by a ground-truth heatmap.
This multi-level supervision allows the network to learn a coarse-to-fine heatmap regression.

\subsubsection{Sensor Model}
In order to use this likelihood volume as a sensor model, it is necessary to map it back to our Markov localisation grid.
To do so, we run a softmax operation along the probability bins, which ensures the sum of all likelihoods for a given \ensuremath{\pose{\mytime}{k}} cell sum to one.
We then take the top \ensuremath{n < N} slices and sum them to obtain a single pose likelihood for the \ensuremath{XY} plane which is then repeated for the \ensuremath{\theta} bins, which we discuss in the following section.
\vspace{-1mm}
\subsection{Convolutional Odometry}\label{conv-odom}
The sensor model introduced in the previous section does not measure the likelihood of the agent's orientation \ensuremath{\theta}.
Instead, we use a novel convolution-based odometry layer as a motion model which estimates \ensuremath{\theta}.
Our model is efficient, so there is no need to artificially limit the update rate other than to guarantee at least one cell of displacement.
This means that, assuming a non-holonomic agent, the motion model can propagate pose likelihoods in a manner that also selects the correct orientation.\looseness=-1

Markov localisation relies on a motion model (equation \ref{motionmodel}) to propagate likelihoods into the correct areas of the grid.
Normally, this is implemented as a shift according to the odometry measurement, followed by diffusion using convolution with a separable Gaussian.
The kernel of this Gaussian is computed based on the odometry's uncertainty.
For a 3-\ac{dof} likelihood grid this is a relatively expensive operation as it would require a set of 3 shifts, and a 3D convolution which would be prohibitively expensive on a CPU.
However, we formulate the odometry kernel as a deep learning layer that enables us to perform an efficient operation on the GPU by mapping the 3 shifts and 3 convolutions into a singe 2D convolution kernel.
By building our sensor and motion model as custom layers on a \ac{cnn}, the entire framework can operate in a single forward pass on one GPU incredibly quickly.
In order to estimate our odometry kernel for 2D convolution, we first look at how the odometry maps into a simple 3D convolution kernel.
The odometry data from any non-holonomic 3-\ac{dof} vehicle can be decomposed as 
\begin{equation}
\mypar{\sca{}{}{\delta\theta}{t}{1}, \sca{}{}{\delta x}{t}{},  \sca{}{}{\delta\theta}{t}{2}} = \easyfunc{odometry\_model}{\odom{\mytime}{}}
\end{equation}
where \ensuremath{\sca{}{}{\delta\theta}{t}{1}} is a rotation followed by a forward translation \ensuremath{\sca{}{}{\delta x}{t}{}} and a final rotation \ensuremath{\sca{}{}{\delta\theta}{t}{2}} \cite{Thrun2006rm}.
Gaussian noise can be applied to each component independently, producing a new set of odometry estimates \ensuremath{\sca{}{}{\hat{\delta}\theta}{t}{1}, \sca{}{}{\hat{\delta} x}{t}{}, \sca{}{}{\hat{\delta}\theta}{t}{2}}.

In order to map this into a 3D kernel, we take this representation and map it into a vector as
\begin{equation}
\vtr{}{}{S}{t}{}=
\begin{pmatrix}
\sca{}{}{s}{x}{} \\
\sca{}{}{s}{y}{}  
\end{pmatrix}
=
\begin{pmatrix}
\sca{}{}{\hat{\delta} x}{t}{} * cos(\sca{}{}{\hat{\delta}\theta}{t}{1}) \\
\sca{}{}{\hat{\delta} x}{t}{} * sin(\sca{}{}{\hat{\delta}\theta}{t}{1}) 
\end{pmatrix}.
\end{equation}
However, this odometry is not directly applicable to every \ensuremath{\theta_c \in \Theta} channel in our likelihood volume.
In order to apply the odometry, we align it such that ``forward'' motion represents the orientation of the \ensuremath{\theta_c} bin.
This is done using a simple 2D rotation matrix defined as
\begin{equation}
\mat{}{}{R}{c}{} = 
\begin{bmatrix}
\cos(\theta_c) & -\sin(\theta_c) \\
\sin(\theta_c) & \cos(\theta_c) \\
\end{bmatrix}
\end{equation}
for  every \ensuremath{\theta_c} in the likelihood volume.
Each of these matrices can then be multiplied with the linear component of the odometry vector
\begin{equation}
\set{}{}{S}{t}{} = \mybra{
\begin{pmatrix}
\mat{}{}{R}{c}{}\vtr{}{}{S}{t}{} \\
\sca{}{}{\delta\theta}{t}{1} + \sca{}{}{\delta\theta}{t}{2}
\end{pmatrix}
\quad \forall \theta_c \in \Theta}
\end{equation}
in order to obtain a set of rotated odometries for each channel.

The set of rotated odometries are directly mapped into a kernel for convolution.
However, it is first necessary map these odometries to the resolution of the likelihood grid, which can be done as
\ensuremath{
\set{}{}{S}{t}{\prime}= \set{}{}{S}{t}{} \odot \vtr{}{}{\lambda}{}{}
}
where \ensuremath{\odot} is element wise multiplication and \ensuremath{\vtr{}{}{\lambda}{}{} = \mytuple{\frac{1}{r_x}, \frac{1}{r_y}, \frac{1}{r_\theta}}} is the resolution of the grid which has been tiled \ensuremath{\Theta} times.

We are not guaranteed that the odometry will be larger than a single cell in the likelihood grid.
For this reason, it is necessary to accumulate odometry measurements over time as
\begin{equation}
\set{}{}{T}{t}{} = \nsum{t}{\set{}{}{S}{i}{\prime}} - \set{}{}{T}{t-1}{\prime}
\end{equation}
where
\ensuremath{
\set{}{}{T}{t}{\prime} = \set{}{}{T}{t}{}-\lfloor\set{}{}{T}{t}{}\rfloor
}
and \ensuremath{\lfloor\set{}{}{T}{t}{}\rfloor} is the odometry applied to the likelihood volume \ensuremath{\poseSet{\mytime}{}}.

\begin{figure}
    \centering
    \includegraphics[width=0.75\linewidth]{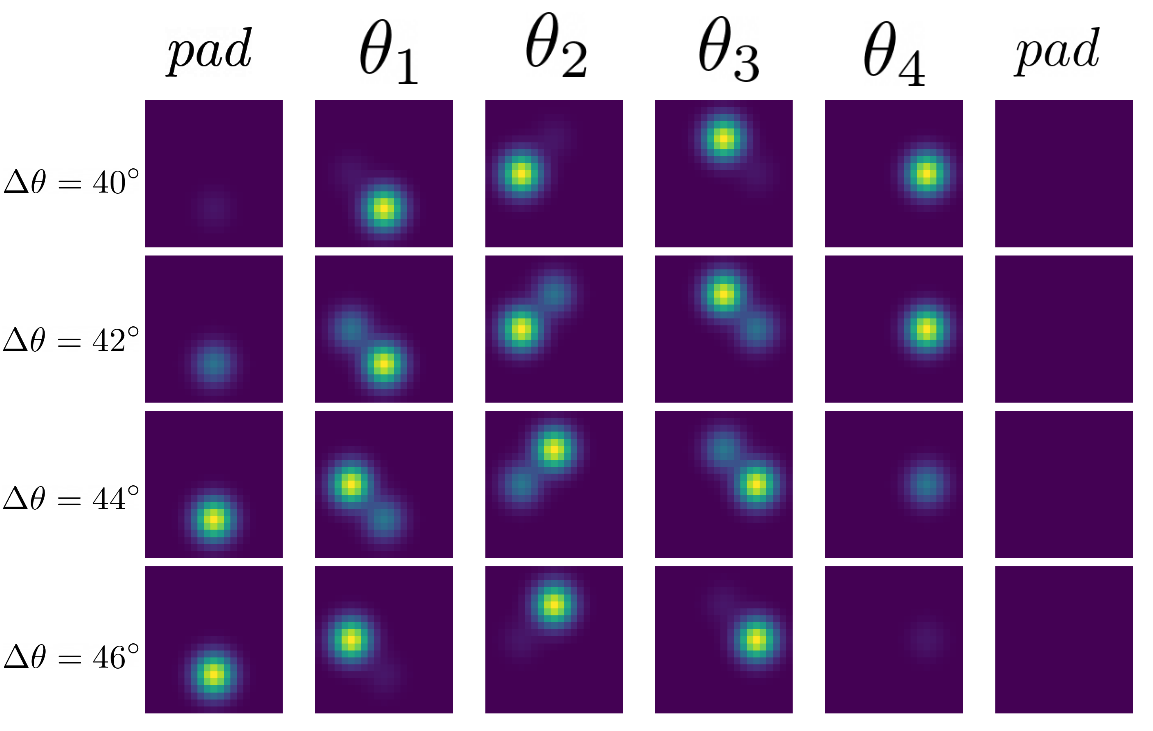}
    \caption{Visualisation of an odometry kernel with pure forward movement. Each column represents a \ensuremath{\theta_c} channel, while each row represents a different odometry measurement for the angle of the vehicle. Notice how each \ensuremath{\theta_c} channel represents the forward motion differently. Similarly, as the rows go down the likelihoods shift between the different theta channels.\looseness=-1}
    \label{fig:kern_step}
    \vspace{-0.45cm}
\end{figure}

\begin{table*}
\centering
\begin{tabular}{l|cccccc}
\textbf{Scene Name} &
\textbf{\begin{tabular}[c]{@{}c@{}}PoseNet \\ \cite{Kendall2015Posenet}\end{tabular}} &
\textbf{\begin{tabular}[c]{@{}c@{}}PoseNet \\ Bayesian\\ \cite{Kendall2015Posenet}\end{tabular}} &
\textbf{\begin{tabular}[c]{@{}c@{}}PoseNet \\ Spatial LSTM\\ \cite{Walch17}\end{tabular}} &
\textbf{\begin{tabular}[c]{@{}c@{}}PoseNet \\ Learn \ensuremath{\beta} \\ \cite{kendall2017geometric}\end{tabular}} &
\textbf{\begin{tabular}[c]{@{}c@{}}PoseNet \\ Geometric \\ \cite{kendall2017geometric}\end{tabular}} &
\textbf{\begin{tabular}[c]{@{}c@{}}Heatmap Regressor \\ (Us)\end{tabular}} \\ \hline

\textbf{GreatCourt}    & -    & -    & -             & 7.00 & 6.83          & \textbf{3.74}  \\
\textbf{KingsCollege}  & 1.66 & 1.74 & 0.99          & 0.99 & \textbf{0.88} & 0.95           \\
\textbf{OldHospital}   & 2.62 & 2.57 & \textbf{1.51} & 2.17 & 3.20          & 2.13           \\
\textbf{ShopFacade}    & 1.41 & 1.25 & 1.18          & 1.05 & 0.88          & \textbf{0.67}  \\
\textbf{StMarysChurch} & 2.45 & 2.11 & 1.52          & 1.49 & 1.57          & \textbf{1.02}  \\
\textbf{Street}        & -    & -    & -             & 20.7 & 20.3          & \textbf{10.21} \\
\end{tabular}
\caption{Median error (m) for Cambridge Landmarks \cite{Kendall2015Posenet}. PoseNet results are from \cite{kendall2017geometric}, where `-' are not reported by authors.}
\label{tab:camb}
\vspace{-0.45cm}
\end{table*}

Applying \ensuremath{\lfloor\set{}{}{T}{t}{}\rfloor} to \ensuremath{\poseSet{\mytime}{}} can be done by converting the rotated odometries into a series of 3D kernels for 2D convolution.
The first step is to map each vector \ensuremath{\vtr{}{}{\tau}{t}{} \in \lfloor\set{}{}{T}{t}{}\rfloor} into a 3D kernel.
To do this we take every element of \ensuremath{\vtr{}{}{\tau}{t}{}} and convert it into a discrete 1D Gaussian kernel of size \ensuremath{\mytuple{k_\theta, k_x, k_y}}.
These kernels are then combined linearly to produce a 3D kernel \ensuremath{\mat{}{}{K}{c}{}} of size \ensuremath{\mytuple{K_\theta, K_x, K_y}}.
This allows us to perform a 2D convolution on a subset of the likelihood grid for each 3D kernel.
This subset consists of the channel the kernel was estimated for as well as the \ensuremath{K_\theta} channels around it.
In order to perform the operation as a single pass of a 2D convolution on GPU hardware, we stack these kernels into a single kernel.
Each 3D kernel is offset so that it is centered on the \ensuremath{\theta_c} channel it corresponds to, giving us a kernel with \ensuremath{\theta_c} Gaussian distributions.
Fig. \ref{fig:kern_step} shows a visualisation of the odometry kernel when \ensuremath{\Theta=4} for a simple forward motion.
Each column represents a \ensuremath{\theta_c} channel (along with padding), while each row represents the rotated odometry for each angle bin. 
Notice how each \ensuremath{\theta_c} channel represents the forward motion differently. 
Similarly, as the rows go down, the likelihoods shift between the different theta channels and respect the circular nature of \ensuremath{\theta} by looping from the last bin to the first. 

While it may seem like these are an expensive set of convolutions, it is important to remember that these are all performed directly on the GPU.
More importantly, our Markov localisation approach is entirely GPU bound. 
Both our sensor and motion model are computed directly on the GPU with no need to ever retrieving the cost volume.
This makes our approach both quick and efficient.
In the following section, we will show that not only is our approach fast but it is also capable of producing state-of-the-art results for localisation.
\vspace{-0.125cm}
\section{Results}

Firstly, we validate the performance of our heatmap regressor by evaluating its performance on the well-established Cambridge Landmarks dataset
\cite{Kendall2015Posenet}.
Secondly, we evaluate on a vehicle navigating a multi-storey carpark.

All the experiments in this section are trained using the same standard architecture for the heatmap regression network: a ResNet50 as the encoder, with 8 decoder blocks to result in a regressed heatmap of \ensuremath{[256 \times 256]} supervised at the last 4 blocks (\ensuremath{M=[32,64,128,256]}). The learning rate was set to \ensuremath{0.0001} with a variable step learning rate and \ensuremath{\gamma=0.5}.

\begin{table*}
\centering
\resizebox{0.9\linewidth}{!}{%
\begin{tabular}{l|cccc|cccc}

&\multicolumn{4}{|c|}{Trajectory 2} &\multicolumn{4}{|c}{Trajectory 3}                                                                                                               \\ 
\textbf{Method}                     & \textbf{RMSE (m)} & \textbf{Mean (m)} & \textbf{Median (m)} & \textbf{Std. Dev. (m)}  & \textbf{RMSE (m)} & \textbf{Mean (m)} & \textbf{Median (m)} & \textbf{Std. Dev. (m)} \\ \hline
PoseNet \cite{kendall2017geometric} & 9.31	            & 5.26	            & 2.66	              & 7.68	              & 14.09             & 6.75              & 3.21                & 12.37                  \\
PoseLSTM \cite{Walch17}             & \underline{9.05}  & 5.04	            & 2.61	              & \underline{7.52}	   & \underline{12.82} & 6.55              & 2.74                & 11.02                  \\
Raw Odometry                        & 12.20	            & 9.23	            & 5.54	              & 7.97	                & 24.86	            & 23.77	            & 24.43	              & \underline{7.25}      \\
Ours (H)                            & 9.74	            & \underline{4.21}	& \textbf{1.36}       & 8.78	               & 16.65	            & 7.14	            & \textbf{1.80}	      & 15.04	               \\
Ours (H+O)                          & \textbf{6.48}	    & \textbf{3.81}	    & \underline{2.45}	  &	\textbf{5.24}          &  \textbf{6.99}	& \textbf{3.88}	    & \underline{2.56}    & \textbf{5.82} \\ 

\end{tabular}%
}
\caption{Average Trajectory Error: PoseNet, PoseLSTM, Ours (Odometry), Ours (Heatmap), Ours (Heatmap and Odometry). Top results are \textbf{Bold}, \underline{underline} second.}
\label{tab:avp_ate}
\vspace{-0.25cm}
\end{table*}

\subsection{Cambridge Landmarks}

The dataset consists of 6 sequences of images captured at different landmarks across the city of Cambridge (UK) using a hand-held camera.
Each sequence consists of anywhere between \ensuremath{300} and \ensuremath{6000} images which are then split between train and test sets.
Ground-truth poses are estimated using \ac{sfm} \cite{Changchang2013} software.
The sensor experiences 6-\ac{dof} motion during capture. 
Furthermore the \ac{sfm} software does not guarantee that there is a well-defined ground plane.
This makes the dataset inherently difficult for our approach, as we only localise in a 3-\ac{dof} space.
Since most of the motion of the sensor is planar, we overcome this limitation by using an \ac{svd} to regress a dominant plane and therefore the height of the sensor.

Table \ref{tab:camb} shows a comparison against several pose-regression networks. 
We show the median pose error on all 3 spatial axis, as reported in \cite{kendall2017geometric}.
It is important to note that this is an unfavourable scenario for our 3-\ac{dof} localisation, as our network does not directly regress the height of the sensor.
We also do not estimate the orientation of the sensor, as the heatmap regressor should be able to cope with the appearance variation.
Regardless, our heatmap regressor is able to outperform several state-of-the-art regression methods in 4 out of 6 scenarios.
We believe there are several reasons for this.
Firstly, a heatmap-based loss provides a more uniform supervision signal than a pose-based loss, as the heatmaps do not scale with the magnitude of the pose space.
Secondly, and most importantly, while we do not strictly enforce multi-modal distributions with our losses, the network is capable of modelling them.
This allows the network to predict uncertain poses without incurring a penalty.
In the KingsCollege scenario, we are only outperformed by PoseNet Geometric \cite{kendall2017geometric}, which makes use of an additional source of supervision: the 3D reconstruction points of the ground truth.
For the OldHospital scenario, we are outperformed by Walch \etal \cite{Walch17} most likely because the self-repeating nature of the architecture is well-suited to the spatial LSTMs they employ.
\vspace{-0.125cm}
\subsection{Multi-Storey Carpark}
One of the main advantages of our approach is the ability to generalise to self-similar environments.
Car-parks are interesting environments for localisation, as they tend to be self-similar within each floor as well as across multiple floors.
By their nature, accurate localisation within a multi-storey car-park requires a multi-modal distribution.

\subsubsection{Data Capture}
\begin{figure}
    \centering
    \begin{tabular}{ccc}
        Input & Ground Truth& Regressed\\
        \includegraphics[width=0.24\linewidth]{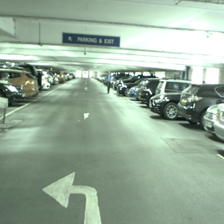}& \hspace{-4mm}
        \includegraphics[width=0.24\linewidth]{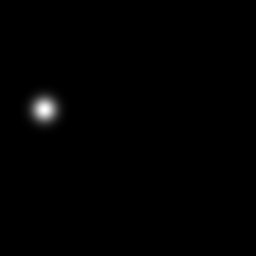}& \hspace{-4mm}
        \includegraphics[width=0.24\linewidth]{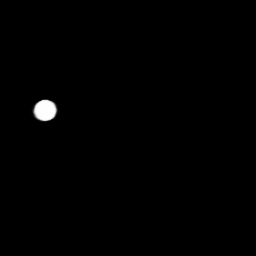}\\
        \newline
        \includegraphics[width=0.24\linewidth]{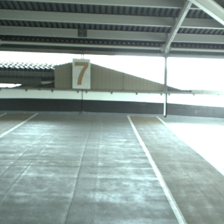}& \hspace{-4mm}
        \includegraphics[width=0.24\linewidth]{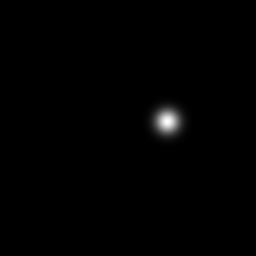}& \hspace{-4mm}
        \includegraphics[width=0.24\linewidth]{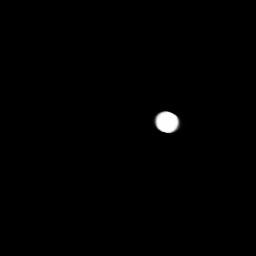}
    \end{tabular}

    \caption{Sample Images from the Multi-Storey Car Park dataset, along with the ground truth heatmap  and output regression.}
    \label{dataset}
    \vspace{-0.55cm}
\end{figure}

This dataset consists of a vehicle driving around a multi-storey car-park.
The vehicle is equipped with 3 front-facing cameras and a 16-beam \ac{lidar}.
It traversed 12 floors, travelling a total of over \ensuremath{6,500 m} and an area of over \ensuremath{7000 m^2}.
The vehicle performed parking manoeuvres such as 3 point turns, bay parking, interacting with traffic, etc.
The \ac{lidar} is used to create ground-truth localisaton data.
The vehicle was driven on two separate days, with two trajectories on the first day and one on the second.
Of the three trajectories, trajectory 1 has \ensuremath{3207} images, 2 has \ensuremath{2860} and 3 has \ensuremath{3922}.
We use trajectory 1 as training data, and reserve trajectories 2 and 3 for testing.
The left column of fig. \ref{dataset} shows sample images from the captured dataset.
This dataset will be released upon publication of this work.

\subsubsection{Heatmap Regression Training}
To train our heatmap regressor we use the pose of the \ac{lidar} along with a calibrated transformation between the sensors.
The estimated pose is then projected to the ground plane, where the heatmaps can be estimated.
The right column of Fig. \ref{dataset} shows two example heatmaps used in training, as well as the regressed heatmaps.

The convolutional odometry consists of a \ensuremath{[72 \times 256 \times 256]} likelihood volume, with a kernel of size \ensuremath{[15 \times 21 \times 21]}.
Using this configuration on an AMD Threadripper 3960X with an Nvidia GeForce GTX 1080 Ti, the heatmap regressor takes \ensuremath{10.701ms} with a standard deviation of \ensuremath{0.0377ms} and the convolutional odometry layer takes \ensuremath{0.253 ms} with a standard deviation of \ensuremath{0.00538 ms}.
In order to simulate odometry, we use the ground-truth poses acquired from the \ac{lidar}. 
For each successive pose, we add noise as described in \cite{Thrun2006rm} (forcing the trajectory to drift) and estimate a set of odometry measurements from the resulting noisy trajectory.

\subsubsection{Quantitative Results} 
We evaluate the performance of our approach against PoseNet \cite{kendall2017geometric} and PoseLSTM \cite{Walch17}, as both represent the state-of-the-art for camera pose regression. 
We compare using the \ac{ate} as established by Sturm \etal \cite{Sturm2012}, which accounts for orientation errors as part of the overall trajectory. 
In Table \ref{tab:avp_ate} it can be seen that our approach outperforms the state-of-the-art by a significant margin. 
This is because our approach has the ability to maintain multiple hypothesis of the pose as the vehicle moves through the car-park. 
By contrast, PoseNet and PoseLSTM are forced to relocalise the camera at every iteration. 
This results in paths that are unreliable. 
Since our approach keeps a likelihood field, failures in the heatmap regression do not directly result in jumps in the pose estimate, therefore smoothing our trajectory and reducing the \ac{ate} error. 
We additionally show the results of the raw odometry measurements and using our heatmap regressor only (with no motion model). 
As it can be seen, the odometry measurements are extremely noisy, resulting in a high \ac{ate}.
The Heatmap Regressor alone outperforms competing approaches but despite the noisy odometry, its addition increase performance further.
\subsubsection{Qualitative Results}
\begin{figure}
    \centering
    \begin{subfigure}[b]{0.49\linewidth}
        \centering
        \includegraphics[width=0.98\linewidth]{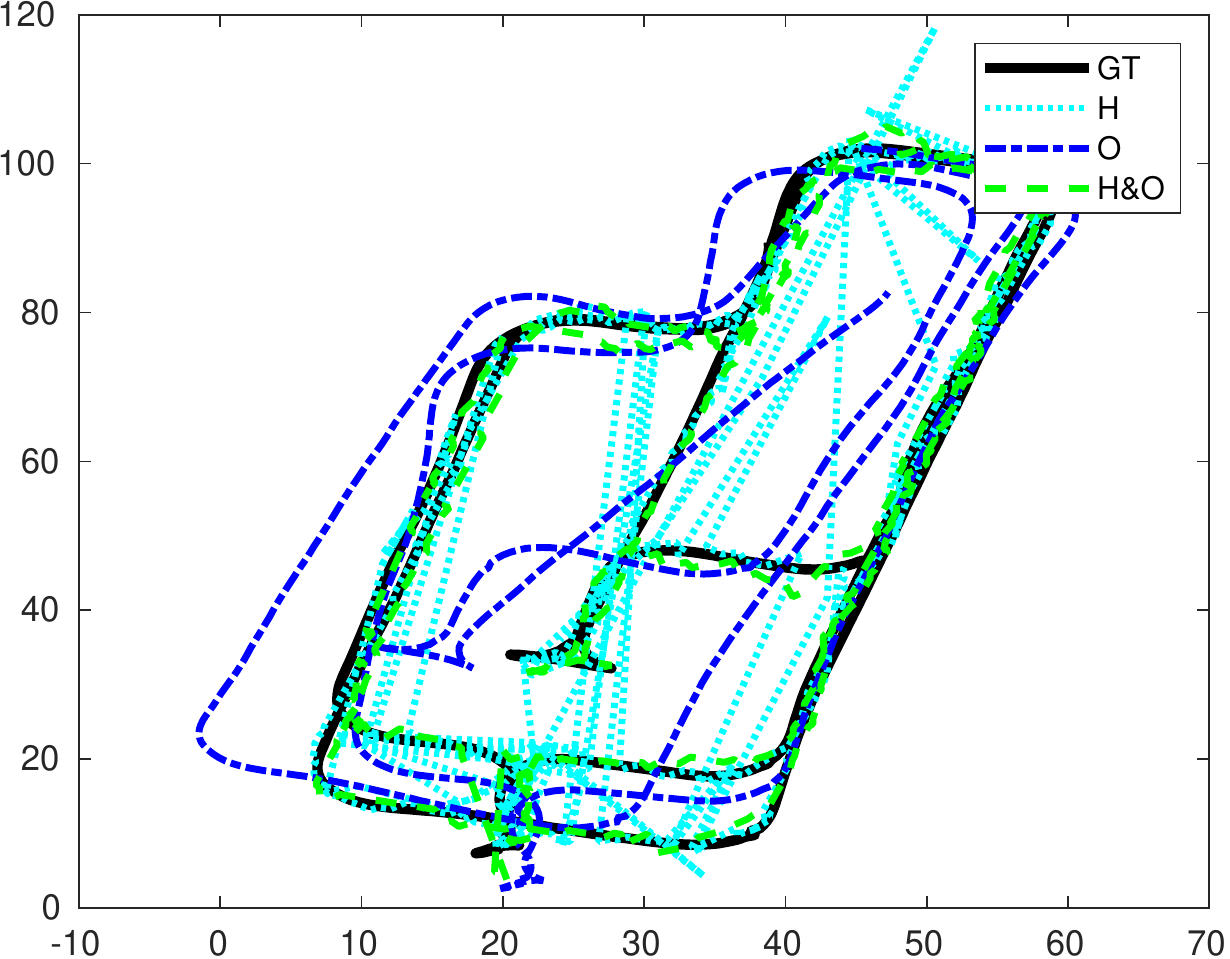}
    \end{subfigure}
    \begin{subfigure}[b]{0.49\linewidth}
        \centering
        \includegraphics[width=0.98\linewidth]{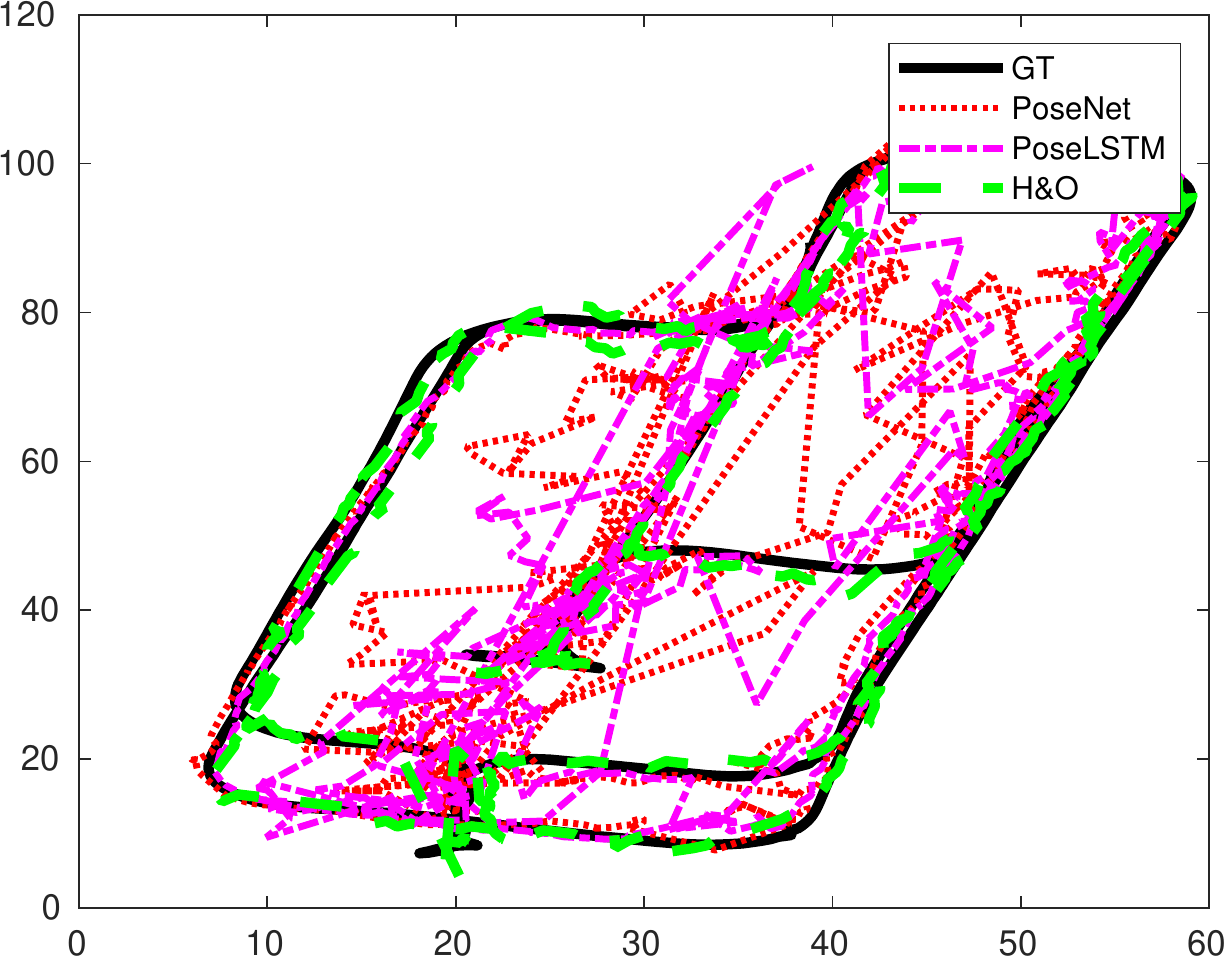}
    \end{subfigure}
    \caption{Trajectories of Ground Truth (GT), PoseNet, PoseLSTM,  Heatmap Regressor (H), Raw Odometry (O) and the Combined Heatmap Regressor \& Convolutional Odometry (H\&O). Note how PoseNet and PoseLSTM do not provide smooth trajectories.} 
    \label{trajs}
    \vspace{-0.55cm}
\end{figure}

In our experience, the numbers presented in tables \ref{tab:camb} and \ref{tab:avp_ate} do not adequately convey the difference in the smoothness of the trajectories.
Fig. \ref{trajs} shows the resulting trajectories from PoseNet, PoseLSTM and our approach, as well as a comparison against the raw odometry and heatmap regressor alone.
For clarity, this is done on the first three floors of the multi-storey carpark.
As it can be seen, our approach is significantly smoother than both PoseNet and PoseLSTM.
It can also be seen that our combined approach is smoother than the independent heatmap regressor, as well as more accurate than the raw odometry measurements.
This is because our likelihood heatmap, convolutional odometry and likelihood grid work together to ensure that poor estimates do not cause jumps in the pose estimate.
Additionally, the ability to represent a multi-modal distribution allows us to make quick corrections when the pose has been incorrectly estimated.\looseness=-1 

\addtolength{\textheight}{-8.7cm}
\vspace{-0.125cm}
\section{Conclusion}
\vspace{-0.125cm}
In summary, we have presented an approach that leverages advances in Deep-learning hardware to perform deep heatmap regression and convolutional odometry, in real time.
Our work operates on commodity GPU hardware and leverages some of the important advances in tensor-based processing by performing all operations directly on the GPU, without the need to transfer data back to the CPU.
Importantly, our approach capitalises on the important probabilistic properties of Markov localisation by exploiting modern parallel GPU technology.\looseness=-1

\bibliographystyle{plain}
\bibliography{MendezICRA2021}   %
\end{document}